\documentclass[lettersize,journal]{IEEEtran}
\usepackage[table]{xcolor}
\usepackage{amsmath,amsfonts}
\usepackage{algorithmic}
\usepackage{algorithm}
\usepackage{array}
\usepackage[caption=false,font=normalsize,labelfont=sf,textfont=sf]{subfig}
\usepackage{textcomp}
\usepackage{stfloats}
\usepackage{url}
\usepackage{verbatim}
\usepackage{graphicx}
\usepackage{cite}
\usepackage{hyperref}       
\usepackage{url}            
\usepackage{booktabs}       
\usepackage{amsfonts}       
\usepackage{nicefrac}       
\usepackage{microtype}      

\usepackage{xcolor}         

\usepackage{enumitem}
\usepackage{arydshln} 
\usepackage{siunitx} 
\usepackage{booktabs} 
\usepackage{multirow} 

\usepackage{graphicx} 
\definecolor{lgray}{gray}{0.85} 
\definecolor{lightgray}{gray}{0.8}

\usepackage{bm}
\usepackage{multirow}
\usepackage{pifont}
\usepackage{makecell} 
\usepackage{amsmath} 
\usepackage{cleveref}
\usepackage{colortbl}  
\usepackage{xcolor}
\usepackage{array}   

\usepackage[rightcaption]{sidecap}
\usepackage[table]{xcolor}
\newcolumntype{g}{>{\columncolor{gray}}c}

\usepackage{placeins}

\definecolor{revise}{HTML}{0071BC}
\def\gain#1{\textcolor{black}{#1}}

\begin{document}

\title{LiteFusion: Taming 3D Object Detectors \\from Vision-Based to Multi-Modal\\ with Minimal Adaptation}

\author{Xiangxuan Ren, Zhongdao Wang, Pin Tang, Guoqing Wang, Jilai Zheng, Chao Ma,~\IEEEmembership{Member,~IEEE}
\IEEEcompsocitemizethanks{
\IEEEcompsocthanksitem This work was supported in part by NSFC (62322113, 62376156), Shanghai Municipal Science and Technology Major Project (2021SHZDZX0102), and the Fundamental Research Funds for the Central Universities. \emph{(Corresponding author: Chao Ma.)}

\IEEEcompsocthanksitem Xiangxuan Ren, Pin Tang, Guoqing Wang, Jilai Zheng and Chao Ma are with the China Ministry of Education (MOE) Key Laboratory of Artificial Intelligence, Artificial Intelligence Institute, Shanghai Jiao Tong University, Shanghai 200240, China (e-mail: \{bunny\_renxiangxuan, pin.tang, guoqing.wang, zhengjilai, chaoma\}@sjtu.edu.cn). Zhongdao Wang is with the Huawei Noah’s Ark Lab (e-mail: wangzhongdao@huawei.com).}}



\maketitle
\begin{abstract}
3D object detection is fundamental for safe and robust intelligent transportation systems. Current multi-modal 3D object detectors often rely on complex architectures and training strategies to achieve higher detection accuracy.  
However, these methods heavily rely on the LiDAR sensor so that they suffer from large performance drops when LiDAR is absent, which compromises the robustness and safety of autonomous systems in practical scenarios.
Moreover, existing multi-modal detectors face difficulties in deployment on diverse hardware platforms, such as NPUs and FPGAs, due to their reliance on 3D sparse convolution operators, which are primarily optimized for NVIDIA GPUs. 
To address these challenges, we reconsider the role of LiDAR in the camera-LiDAR fusion paradigm and introduce a novel multi-modal 3D detector, LiteFusion. Instead of treating LiDAR point clouds as an independent modality with a separate feature extraction backbone, LiteFusion utilizes LiDAR data as a complementary source of geometric information to enhance camera-based detection. This straightforward approach completely eliminates the reliance on a 3D backbone, making the method highly deployment-friendly.
Specifically, LiteFusion integrates complementary features from LiDAR points into image features within a quaternion space, where the orthogonal constraints are well-preserved during network training. This helps model domain-specific relations across modalities, yielding a compact cross-modal embedding. 
Experiments on the nuScenes dataset show that LiteFusion improves the baseline vision-based detector by $+$20.4\% mAP and $+$19.7\% NDS with a minimal increase in parameters (1.1\%) and achieves comparable performance to previous multi-modal detectors without using dedicated LiDAR encoders. 
Notably, even in the absence of LiDAR input, LiteFusion maintains strong results (46.6\% NDS and 39.2\% mAP), highlighting its favorable robustness and effectiveness across diverse fusion paradigms and deployment scenarios. 
\end{abstract}


\section{Introduction}
\label{sec:intro}

\IEEEPARstart{R}{eliable} 3D object detection is essential for intelligent transportation systems, especially in autonomous vehicles, where understanding the environment ensures safe navigation and decision making~\cite{song2024robustness, aung2024review}. Accurate 3D object detection depends significantly on integrating complementary information from both cameras and LiDAR.
Conventional multi-modal 3D object detectors often process each modality separately, necessitating distinct processing pipelines for cameras and LiDAR.
To pursue higher performance, recent multi-modal detectors have adopted more complex fusion architectures and training strategies. As shown in Fig.~\ref{fig:introduction}(a), integrating camera and LiDAR data often involves independent feature extraction followed by intricate fusion processes, such as bidirectional coordinate projections, additional depth estimation models, and sophisticated fusion architectures. Despite their potential, this dual-stream fusion approach faces several challenges:

\begin{figure*}[tb]
  \centering
  \includegraphics[width=0.95\linewidth]{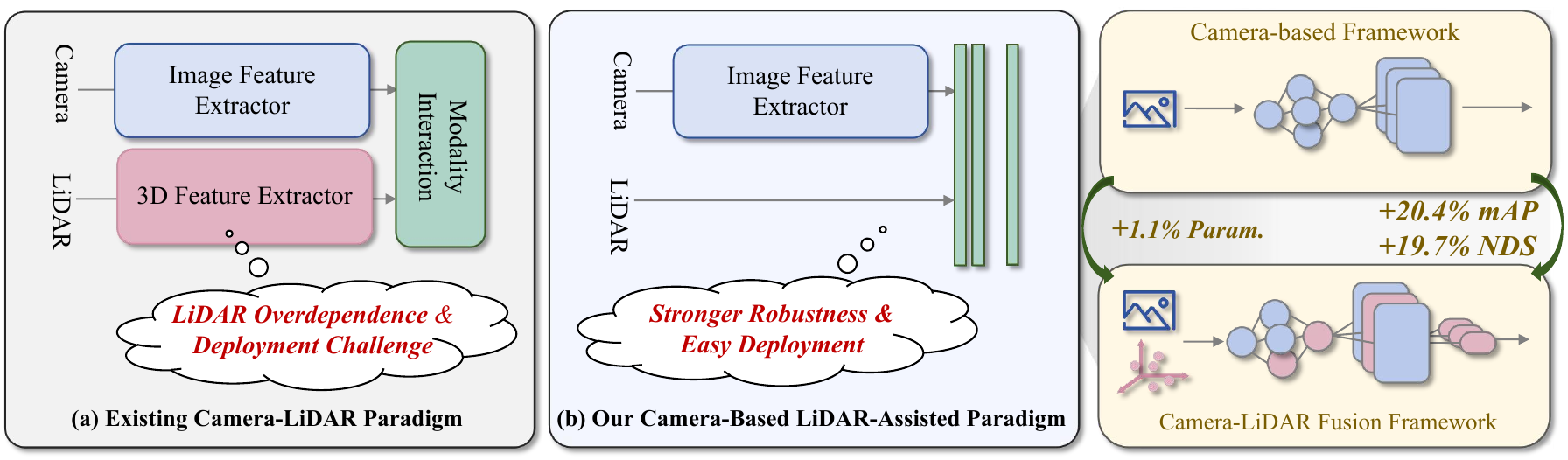}
  \caption{\textbf{Illustration of the camera-LiDAR fusion paradigms.} 
  Existing multi-sensor fusion methods often rely on large 2D and 3D backbones, complex modality interactions, and elaborate training stages. In contrast, our approach simplifies the pipeline by using a progressive modality interaction strategy, eliminating the need for a 3D feature extractor. By adding only 1.1\% more parameters to the camera-based framework, we achieve a significant boost in 3D perception performance, with improvements of +20.4\% in mAP and +19.7\% in NDS.
  }

  \label{fig:introduction}
\end{figure*}

\noindent \textbf{i) Excessive reliance on LiDAR.} 
We observe that the performance of multi-modal detectors degrades significantly when LiDAR points are missing, e.g., from 71.4\% NDS to 7.1\% NDS in BEVFusion\cite{bevfusion}, falling well below that of vision-based detectors.
We reason that existing cross-modal 3D detectors favor the intuitive spatial geometry information provided by LiDAR while not paying enough attention to the important texture details inherent in image data. This excessive reliance on LiDAR compromises the robustness and reliability of multi-modal detectors, particularly in scenarios involving sensor failures.

\noindent \textbf{ii) Deployment challenge.} 
Most separate LiDAR encoders~\cite{second,voxelnet,voxelnext,dsvt,vsc} are built upon sparse 3D convolution operators, which require customized implementation and optimization to perform well across various hardware platforms. This limitation can hinder their use in real production environments. For instance, on common on-device inference platforms such as FPGAs (e.g., Xilinx), NPUs (e.g., Huawei Ascend, Horizon Zhengcheng), or GPUs from other brands like AMD, sparse convolutions are not optimized to the same extent as they are on NVIDIA GPUs.

\noindent \textbf{iii) Data scaling challenge.} 
The emergence of large vision foundation models highlights the scalability of image-based methods, underscoring the importance of image modality in multi-modal approaches. On the other hand, the expansion of multi-modal data is limited by the different deployment strategies and configuration variations among LiDAR sensors, which restricts the scalability of dual-stream fusion models.

To overcome the aforementioned limitations, we question whether the existing fusion scheme with separate modality encoders truly offers a sustainable advantage. 
In response, we propose a novel \textbf{camera-based LiDAR-assisted} fusion scheme by integrating the dual-stream network architecture into a unified model to manage both modalities, thus eliminating the complex point cloud backbone, as shown in Fig.~\ref{fig:introduction}(b). To achieve this, it is natural to start from a camera-based detector and tame it to a multi-modal detector with minimal structural adjustments.
However, directly incorporating raw LiDAR points into the camera-based framework brings a negligible performance gain (only $+0.9\%$ mAP, $+2.7\%$ NDS on nuScenes~\cite{nuscenes}), due to the significant domain discrepancy through the network.
Therefore, the principal challenge lies in developing a straightforward strategy to mitigate the domain gap between 3D LiDAR geometry and 2D visual information without requiring an additional independent 3D feature extractor or major modifications to existing detectors. 

We further present a novel approach, dubbed \textbf{LiteFusion}. Based on the camera-based LiDAR-assisted fusion scheme, LiteFusion regards LiDAR data as a supplementary source of 3D geometric information to enhance and complement camera-based detection, rather than a separate modality requiring a conventional feature backbone. 
A significant advancement is that we introduce the quaternion space for cross-modal fusion and interaction, where LiteFusion learns to map 3D geometry information from LiDAR data to 2D images.
We interpret that modality fusion is not a simple \textit{alignment} of two modalities; rather, it involves reinforcing the \textit{unique modality-specific information} across modalities to fully utilize both modalities.
By compactly encoding the inter-modal relationships between domain-specific image features and point cloud features within a hypercomplex hidden space, the quaternion representation efficiently encodes denser information with fewer parameters. Besides, the orthogonal constraints within the quaternion layer make sufficient use of geometric information from LiDAR data. This not only improves the camera-based detection process by integrating essential geometry context and depth cues from the point cloud, but also enhances robustness against variations in input data when LiDAR data is absent.
Moreover, unlike the conventional one-step fusion scheme~\cite{bevfusion, transfusion, cmt}, LiteFusion progressively incorporates the geometric information in the point cloud into the image-based detectors to achieve additional performance gains. 
With the above design, LiteFusion has the following merits:

    \noindent\textbf{i) Balanced modality utilization:}  
    LiteFusion is engineered to optimally balance the use of both visual and LiDAR data, regardless of the availability of LiDAR data. 
    This configuration offers enhanced reliability in real situations where LiDAR inputs may be compromised or unavailable.

    \noindent \textbf{ii) Deployment-friendly integration:}
    LiteFusion relies solely on standard and well-optimized operators like \texttt{conv2d}, without requiring 3D sparse convolution operators. Thus, it is easily deployable on any platform beyond NVIDIA GPUs. 
    Notably, we have successfully deployed LiteFusion on the Huawei Atlas 910B NPU without customized optimization.

    \noindent \textbf{iii) Seamless detector conversion:} 
    LiteFusion inherits parameters from vision-only model and integrates LiDAR data as geometric cues with little increase in the number of parameters (i.e., only +1\%), thereby promoting scalability to larger vision datasets or better pre-trained vision models.
    
In summary, we make the following contributions:
\begin{itemize}
\item We propose LiteFusion, a unified framework that simplifies the dual-stream network architecture by eliminating the need for complex 3D feature extraction processes, enabling deployment-friendly and robust camera-LiDAR data integration.

\item We introduce quaternion space embedding for modality fusion in cross-modal adaptation, which offers an innovative perspective to effectively map 3D LiDAR geometry onto 2D images, thus facilitating compact encoding of inter-modal relationships with fewer parameters.

\item   
Extensive experiments demonstrate that LiteFusion delivers performance on par with state-of-the-art methods without requiring a 3D sparse convolution extractor and performs favorably in scenarios where LiDAR data is absent.
\end{itemize}

\section{Related Work}

\subsection{3D Object Detection}

Recent progress in 3D perception in autonomous driving has been notably advanced.
To detect 3D objects from 2D images, depth-based methods~\cite{lss, bevdepth, bevdet,ealss,sparse,vsc} focus on estimating precise depth distributions to lift 2D image characteristics to 3D frustum space.
However, this paradigm is sensitive to depth estimation, where minor estimation errors often cause major mismatches in 2D-3D mapping and yield heavy detection degradation. 
Query-based methods~\cite{detr3d, bevformer,sparsebev,uni3detr, zhang2024transfusion} treat 3D object detection as a set prediction problem, significantly simplifying the prediction process by narrowing the search space to a fixed set of learnable queries.
Benefiting from the flexibility of this paradigm, encoding long time-series information into the feature encoder leads to increasingly high performance~\cite{bevstereo, detr4d, lin2023sparse4d, tbpformer, solofusion}.
Compared to vision-based methods, training fusion models for high performance is complex and often requires a multi-stage process, which increases training costs. This typically involves training the image and LiDAR branches separately, followed by combined training of the fusion module.  
For processing point clouds, most detectors convert point clouds into regular grid structures like voxels\cite{voxelnet, second,sparsefusion}, pillars\cite{pointpillars,pillar,center}, or range images\cite{rangedet}. 
However, the sparse convolution layers typically used for extracting 3D features from point clouds require specialized hardware or software optimization to function effectively.
Our method aims to eliminate the LiDAR feature extraction component altogether, opting instead for a more efficient camera-based LiDAR-assisted framework to address the challenges of LiDAR data encoding.

\subsection{Camera-LiDAR Fusion Strategies}
The rich texture and color information provided by cameras enhances the 3D geometric data obtained from LiDAR, resulting in a more comprehensive understanding of the environment. 
Existing fusion schemes mainly fall into two categories: explicit geometric projection fusion and deep feature fusion. Explicit geometric projection fusion~\cite{pointaugmenting, pointpainting, mv3d} leverages the inherent geometric properties of sensor data to integrate data effectively. However, its success depends heavily on the accuracy of sensor calibration and the precision of geometric transformations. In contrast, the deep feature fusion scheme~\cite{bevfusion, bevfusion1, uni3d, supfusion,objectfusion} abstracts sensor data into a high-dimensional hidden feature space for integration, offering a more flexible and potentially more powerful approach to data fusion.
Within the transformer framework, proposal-based methods~\cite{futr3d, cmt, transfusion, deepinteract} aggregate modal features using queries at the detection head. By utilizing specially designed fusion networks~\cite{logonet, msmdfusion, unitr}, these models can perform learnable feature alignment within an implicit space. 
Lacking clear physical guidelines, fusion models risk the loss of important information from one sensor while over-relying on another. Moreover, relying on modality-specific encoders can complicate the practical deployment of advanced models in real scenarios. 
Unlike existing approaches, our method directly maps 3D geometric information from LiDAR data onto 2D image features within the quaternion space, providing a robust and deployment-friendly solution for camera-LiDAR fusion.

\begin{figure*}[!t]
  \centering
  \includegraphics[width=0.95\linewidth]{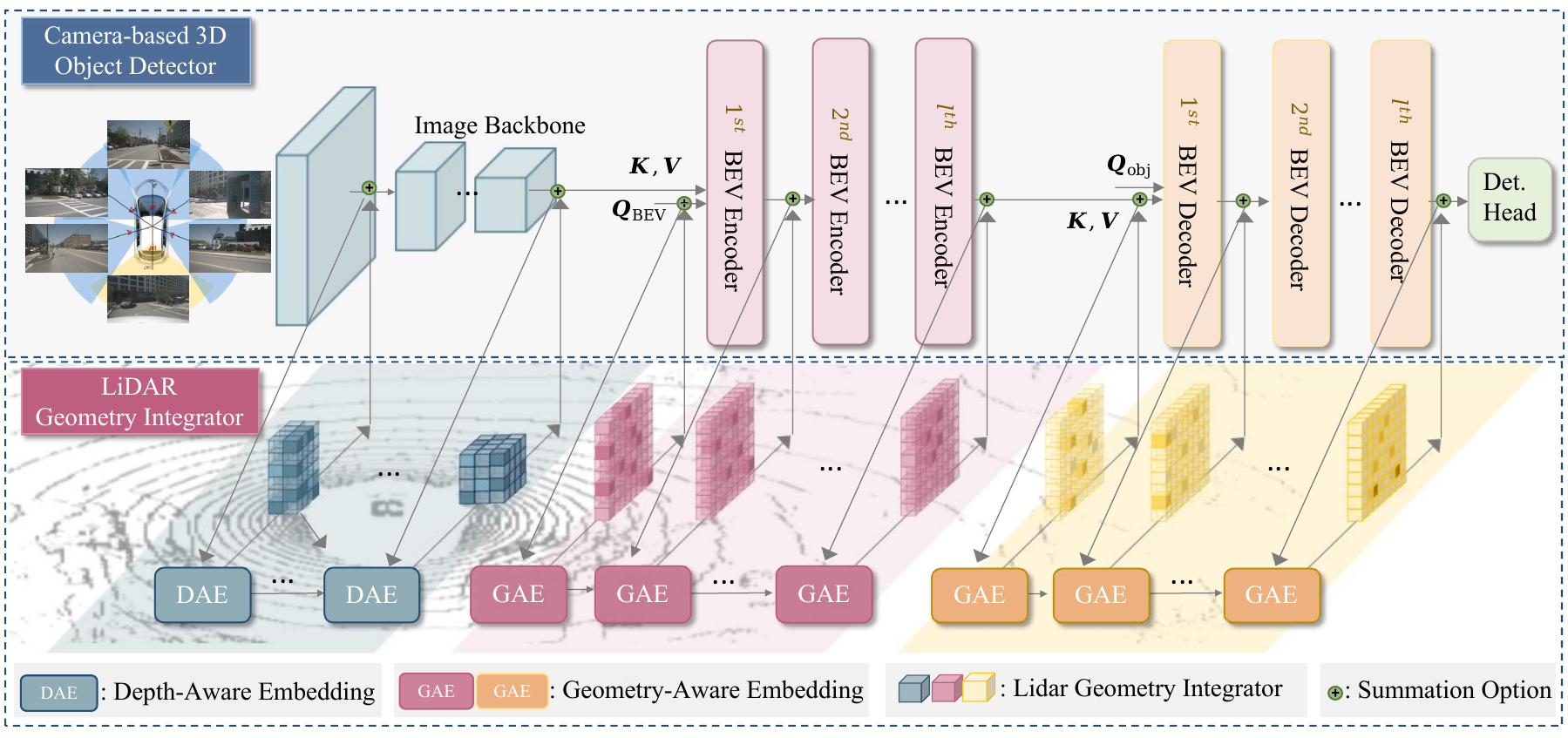}
  \caption{\textbf{Overall framework of the proposed approach.} 
  LiteFusion leverages the camera-based LiDAR-assisted fusion scheme, taming a camera-based 3D object detector to a multi-modal detector via the devised LiDAR geometry integrator, where the proposed DAE and GAE modules efficiently generate LiDAR-derived geometric information from the PV and BEV perspectives separately. These LiDAR features are hierarchically forwarded to the vision-based detector to enhance the image feature, progressively bolstering the 3D spatial awareness and performance of the detector.
  }

  \label{fig:overview}
\end{figure*}

\section{Methodology}

This work investigates the foundational relationship between two critical modalities in 3D object detection: images and point clouds. 
We introduce LiteFusion, a framework designed to enhance image-based 3D perception by leveraging LiDAR for 3D spatial aggregation, as illustrated in Fig.~\ref{fig:overview}. LiteFusion redefines the role of LiDAR by streamlining its integration into the image feature space, resulting in a lightweight paradigm to boost the performance of 3D detectors.

\subsection{Preliminary and Notation} 

\noindent\textbf{Problem setting.} 
We address the challenge of adapting a vision-based 3D detection framework $\bm{D}_{\text{cam}}:{\bm{I}_{\text{cam}}} \rightarrow \bm{Y}$, which infers 3D bounding boxes $\bm{Y}$ from camera inputs $\bm{I}_{\text{cam}}$, to process point clouds $\bm{I}_{\text{LiD}}$ with minimal modifications.
This involves seamlessly integrating the visual context from cameras with the geometric information from LiDAR. We refer to this task as camera-based LiDAR-assisted 3D object detection, with a focus on achieving a smooth transition from camera-only to multi-modal detectors with minimal modifications.

\noindent\textbf{Foundation model.} 
Camera-based 3D object detectors generally consist of several key components: an image backbone, a Bird's Eye View (BEV) encoder, a BEV decoder, and a detection head. We utilize BEVFormer as our primary camera-based 3D object detector. During inference, multi-view images are input into the image backbone (such as ResNet~\cite{resnet101}, Swin-T~\cite{swin}, and ConvNext~\cite{convnext}), which generates features \( \bm{F}_{\text{img}} = \{\bm{F}_i\}_{i=1}^{N_{\text{cam}}} \) from different camera viewpoints. Here, \( \bm{F}_i \) represents the features from the \( i \)-th view, and \( N_{\text{cam}} \) indicates the total number of camera views.
Within each layer of the BEV encoder, BEV queries \( \bm{Q}_{\text{BEV}} \) simultaneously leverage temporal data from previous BEV features and spatial details from the multi-view features. This process employs a combination of self-attention and cross-attention mechanisms. After passing through \( N_{\text{e}} \) encoder layers, the model generates unified BEV features, which are subsequently processed by the BEV decoder and passed to the detection head for final outputs.

\noindent\textbf{Quaternion algebra.} In the domain of quaternion algebra, a quaternion ${Q}$ represents an extension of complex numbers into a four-dimensional space:
\begin{align}
Q = r1 + xi + yj + zk,
\end{align}
where $r, x, y, z \in \mathbb{R}$. The elements $1$, $i$, $j$, and $k$ constitute the basis units of quaternion. In this representation, $r$ serves as the real component, while the combination $xi + yj + zk$, obeying the relations $i^2 = j^2 = k^2 = ijk = -1$, forms the imaginary or vector part of the quaternion. 
The representation of quaternion ${Q}$ can also be reinterpreted in a matrix form for computational efficiency, given by:
\begin{align}
Q_{\text{matrix}} = 
\begin{bmatrix}
   r & -x & -y & -z \\
   x &  r & -z &  y \\
   y &  z &  r & -x \\
   z & -y &  x &  r 
\end{bmatrix},
\end{align}
Besides, we build upon prior works~\cite{quaternion1,quaternion2,quaternion3} that extend quaternion operators to neural networks by representing inputs, weights, outputs, and biases as quaternions following quaternion algebra. The quaternion’s structure enables the compact encoding of complex spatial transformations, crucial for 3D rotations and orientation tracking, thereby benefiting the encoding of spatial geometric information in point clouds~\cite{zhouquatenion,lidarquaternion}.

\subsection{Overall Progressive Response Framework} 
To unify the dual-stream cross-modal network into a single-stream model, LiteFusion employs a progressive response framework, as illustrated in Fig.~\ref{fig:overview}. This framework replaces the 3D feature extraction backbone with a series of LiDAR geometry integrators that efficiently incorporate 3D geometric information from the point cloud at each processing stage.

We specifically designed two integration mechanisms to enhance camera-based detection workflows: the Depth-Aware Embedding (DAE) module for the image backbone and the Geometry-Aware Embedding (GAE) module for the BEV encoder and decoder. 
Collectively, we refer to these two modules as the LiDAR Geometry Integrator (LGI). 
When LiteFusion receives point cloud input \(\bm{I}_{\text{LiD}}\), it first projects this data into both perspective view (PV) and BEV formats, generating \(\bm{M}_{\text{depth}}\) and \(\bm{M}_{\text{BEV}}\), respectively. Next, \(\bm{M}_{\text{depth}}\) and \(\bm{M}_{\text{BEV}}\) are processed through the DAE and GAE modules and produce learnable LiDAR geometry features to complement image features. 
This yields LiDAR geometry features with 3D spatial cues in a layer-by-layer manner:
\begin{equation}
   \bm{C}_{*}^{s} = \mathcal{G}^s(\bm{F}_{*}^{s}, \bm{C}_{*}^{s-1}), \\
   \quad s=1,2,\ldots, l
\end{equation}
where \(\mathcal{G}^s\) represents the \(s\)-th LiDAR geometry integrator, 
\(\bm{F}_{*}^{s}\) refers to the image feature at the \(s\)-th stage, and \(\bm{C}_{*}^{s-1}\) denotes the spatial geometry feature generated in the previous phase. 
Specifically, \(\bm{C}_\text{depth}^{0}\) is initialized with \(\bm{M}_{\text{depth}}\) for DAE, while \(\bm{C}_\text{geo}^{0}\) is set to \(\bm{M}_{\text{BEV}}\) for GAE.
The learned features \(\bm{C}_{*}^{s}\) progressively integrate the geometric information along with each forward propagation, thereby largely enhancing the camera features at various stages.

This layer-by-layer integration of point cloud data ensures that the network makes full use of the raw LiDAR signals. Geometrical information is gradually integrated into camera features without significant parameter increases.

\subsection{LiDAR Geometry Integrator}

The workflow of the LiDAR geometry integrator begins with the DAE estimating depth information from images. Then, equipped with GAE, the BEV encoder and decoder progressively aggregate the depth-aware and geometry-enhanced features into the BEV map. 
These enhanced features are then fused back into the image feature space, ensuring that both depth and spatial geometric information contribute directly to refining the final detection results. 

\subsubsection{Depth-Aware Embedding}
\begin{figure*}[tb]
  \centering
  \includegraphics[width=0.95\linewidth]{figures/DAE.pdf}
  \caption{\textbf{Detailed design of the proposed DAE.} (a) DAE aligns the depth-aware feature dimensions with camera features, followed by a convolution operation to streamline the latent space dimensions. The two data streams are then concatenated and projected into suprasphere space using a quaternion layer for effective fused feature representation. Key computational steps of the quaternion network are illustrated in (b). Finally, the enhanced features are mapped back to their original dimensions.}
  \label{fig:DAE}
\end{figure*}

The DAE module aims to model the cross-modal relationship within a unified hidden space. This involves an interactive and complementary exchange of domain-specific information. We empirically find that directly incorporating unprocessed raw point clouds into a 2D backbone does not work well. We attribute this issue to two main factors: \textbf{i)} the domain gap that arises from using image feature extraction backbones pre-trained on datasets such as ImageNet~\cite{imagenet}, and \textbf{ii)} the inherent loss of information that occurs when point clouds are integrated into the image processing stream without appropriate constraints.

To effectively bridge the gap between these two modalities, our work focuses on optimizing feature mapping to reduce redundancy and enhance critical geometric information. 
We propose a \textbf{Quaternion Feature Adaptation (Qua-FA)} module that employs quaternion layers to model orthogonal relationships and explore inter- and intra-correlations within the quaternion hidden space. This approach preserves unique modality-specific details across modalities and effectively captures these correlations within the hyper-complex number space.

As shown in Fig. \ref{fig:DAE}(a), DAE follows a two-step methodology. First, it maps the features of both modalities to a lower-dimensional space to align and integrate them. Then, it upgrades the dimensions to generate depth-aware features. For image feature extraction, the DAE processes the depth map \(\bm{M}_\text{depth}\) alongside high-dimensional image features, augmenting the depth-awareness capability of 2D feature extraction backbones. Specifically, given the \(s\)-th stage of the image backbone output features \(\bm{F}_{\text{img}}^{s} \in \mathbb{R}^{H_s \times W_s \times C_s}\) and the depth-aware features generated in the previous stage \(\bm{C}^{s-1}_{\text{depth}} \in \mathbb{R}^{H'_s \times W'_s \times C'_s}\), DAE first performs maximum padding and interpolation to unify the shape of \(\bm{C}^{s-1}_{\text{depth}}\) with the image features, resulting in \(\widetilde{\bm{C}}^{s-1}_{\text{depth}} \in \mathbb{R}^{H_s \times W_s \times C'_s}\).
Subsequently, DAE conducts channel reduction to obtain a compact representation using \(1 \times 1\) convolutions \(g_1(\cdot)\) and \(g_2(\cdot)\):
\begin{equation}
\hat{\bm{F}}_{\text{img}} = g_1(\bm{F}_{\text{img}}^{s}),\;  \hat{\bm{C}}_{\text{depth}} = g_2(\widetilde{\bm{C}}^{s-1}_{\text{depth}})
\end{equation}

To effectively model the relationship between domain-specific vision features \( \hat{\bm{F}}_{\text{img}} \) and the learnable depth categories \( \hat{\bm{C}}_{\text{depth}} \), we employ quaternion feature adaptation to impose orthogonal constraints within the quaternion hidden space. Specifically, the two modalities are concatenated and fed into a quaternion network, where they are internally separated and positioned on the real and the first imaginary axis. This process is mathematically expressed as:
\begin{equation}
\bm{Q}_h = \hat{\bm{F}}_{\text{img}}r + \hat{\bm{C}}_{\text{depth}}i + 0j + 0k
\end{equation}
where \( r, i, j, k \) are the basis units, with \( r \) representing the real part and \( xi + yj + zk \) the vector part. A quaternion mapping layer function, denoted as \( \mathcal{Q}ua(\cdot) \), can be formally represented by:
\begin{equation}
    \mathcal{Q}ua(\bm{Q}_h) = \alpha(\bm{W} \odot \bm{Q}_h)
\end{equation}
where \( \bm{W} \) represents the learnable parameters within the quaternion layer, and \( \odot \) denotes the Hamilton product, as shown in Fig.~\ref{fig:DAE}(b). 
Specifically, the Hamilton product transforms elements in hyper-complex space by grouping input features along the four orthogonal axes, namely \( \{r, i, j, k\} \). For each axis, the corresponding components of the input feature (\(Q_r^{\text{in}}, Q_i^{\text{in}}, Q_j^{\text{in}}, Q_k^{\text{in}} \)) are orthogonally weighted and updated by their respective weights~\cite{quaternion2, quaternion3}, represented as \( W_r, W_i, W_j, W_k \). 
This product enables the quaternion mapping layer to capture and model intrinsic latent connections among features distributed across the four axes.
The activation function \( \alpha(\cdot) \) applied to a quaternion \( \bm{Q_h} \) is defined as:
\begin{equation}
\alpha(\bm{Q_h}) = f(r) + f(x)i + f(y)j + f(z)k,
\end{equation}
where \( f(\cdot) \) is a function applied element-wise to the real (\( r \)) and imaginary  (\( x \), \( y \), \( z \)) components of the quaternion. In the quaternion layer, the weights are obtained from a suprasphere and are initialized as orthogonal weights. 
Finally, the aligned features are mapped back to their original dimensionality, producing the final learned depth-aware feature \( \bm{C}_{\text{depth}}^s \). The initial depth-aware feature \( \bm{M}_{\text{depth}} \) is supervised by the depth map derived from the point cloud. 

In DAE, the orthogonal inter-modal relationship between domain-specific vision texture features and LiDAR geometry embeddings is well-mined within the quaternion hidden space, and the generalized spatial embeddings are projected into the specialized domain. This process integrates geometric information from point clouds into image features, guiding the feature extraction process to leverage depth cues and enhance spatial geometry reasoning by embedding 3D geometric priors into the 2D image domain.

\subsubsection{Geometry-Aware Embedding}

\begin{figure}[!t]
  \centering
  \includegraphics[width=9.0cm]{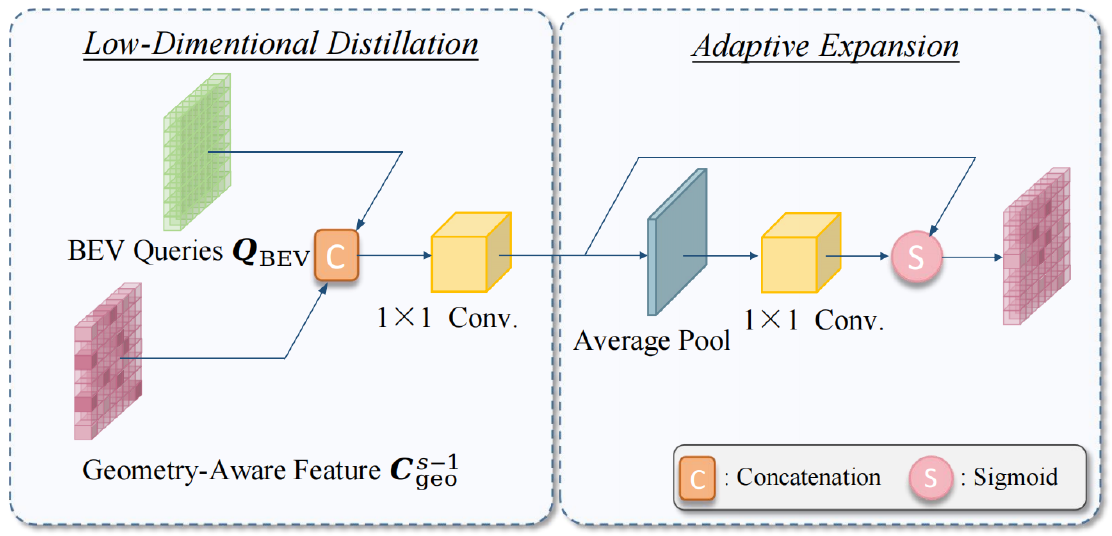}
  \caption{\textbf{Detailed design of the proposed GAE}. The input streams are combined through the concatenation operation, followed by a convolution operation to distill them into a lower-dimensional latent space, after which the adaptive expansion operation is used to spatially align the data.}
  \label{fig:GAE}
\end{figure}

After the image feature extraction process, the BEV encoder employs a set of learnable queries \(\bm{Q}_\text{BEV} \) to spatially map the high-dimensional features encoded by the image backbone to the dense BEV map. This dense map is subsequently fed into the BEV decoder for final detection. 
To effectively integrate LiDAR geometric information into the BEV map, we propose a dynamic embedding module called Geometry-Aware Embedding (GAE). This module is designed for both the BEV encoder and decoder, utilizing the same architecture for efficient feature transformation and mapping.

As shown in Fig.~\ref{fig:GAE}, given the BEV query \(\bm{Q}_{\text{BEV}}\) and the previously obtained LiDAR geometry-aware features \(\bm{C}^{s-1}_{\text{geo}} \in \mathbb{R}^{H_b \times W_b \times C_b}\), we project both data flows into a reduced dimension \(\hat{c}\) using the projection function \(\hat{g}(\cdot)\):
\begin{equation}
\bm{F_{\text{align}}} = \hat{g}([\bm{Q}_\text{BEV}, \bm{C}_{\text{geo}}^{s-1}]), 
\end{equation}
Following the alignment process, an adaptive expansion operation selectively refines \(\bm{Q}_\text{BEV}\) by incorporating \(\bm{F}_{\text{align}}\), thereby embedding 3D geometric cues into the BEV queries. This module can be defined as:
\begin{equation}
\bm{C}_{\text{geo}}^s = \delta(\bm{W}_t \cdot f_{\text{avg}}(\bm{F}_{\text{align}})) \cdot \bm{F}_{\text{align}},
\end{equation}
where \(\bm{W}_t\) denotes the linear transformation matrix, \(f_{\text{avg}}\) is the global average pooling operation, and \(\delta\) is the Sigmoid function.
Notably, for the first layer, we start from the BEV map \(\bm{M}_{\text{BEV}}\), obtained through straightforward voxelization of the point cloud and subsequent compression along the z-axis. GAE then applies an embedding layer to align its dimensions with \(\bm{Q}_{\text{BEV}}\), producing a new representation \(\bm{M}_{\text{BEV}}{\prime}\), which serves as the initial input \(\bm{C}_{\text{geo}}^{0}\) for GAE.

GAE does not utilize quaternion feature adaptation due to the homogeneous nature of its input data, where inter-modal differences are minimal. In contrast, DAE handles heterogeneous data with significant modal discrepancies, necessitating quaternions for their robust orthogonal properties. 
During the encoding process, the queries \(\bm{Q}_\text{BEV}\) are refined at each layer using a newly generated geometry-aware feature \(\bm{C}^{s}_{\text{geo}}\), which progressively alters the composition of the BEV map. This layer-wise adjustment preserves and enhances important geometry features from the point cloud data, enriching the model's understanding of 3D spaces. 

\section{Experiments}
\label{sec:experiments}

\begin{table*}[!t]
  \centering
  \caption{\textbf{Result comparison on nuScenes val set.} 
  The setting of whether to use the 3D backbone is listed separately to emphasize the impact of point feature extraction on model performance. 
  LiteFusion achieves excellent performance without the 3D backbone, while existing methods suffer severe performance drops when the 3D backbone is removed. LiteFusion also presents the best robustness when LiDAR data is missing.
}
  \label{tab:main}
  \small 
  \setlength{\tabcolsep}{18pt} 
  \renewcommand{\arraystretch}{1} 
  \resizebox{\textwidth}{!}{
   \begin{tabular}{l
   clcccc}
    \toprule[1pt]
    \multirow{2}{*}{\textbf{Methods}} & \multirow{2}{*}{\shortstack{\textbf{3D}\\ \textbf{Backbone}} }& \multirow{2}{*}{\shortstack{\textbf{2D}\\ \textbf{Backbone}} }& \multicolumn{2}{c}{\textbf{Camera$+$LiDAR}} & \multicolumn{2}{c}{\textbf{Missing LiDAR}}  \\
    \cmidrule(lr){4-5} \cmidrule(lr){6-7}
    & & &{\textbf{NDS$\uparrow$}} & {\textbf{mAP$\uparrow$}} & {\textbf{NDS$\uparrow$}} & {\textbf{mAP$\uparrow$}}\\
    \midrule
    \midrule
    PointPainting~\cite{pointpainting} & \ding{52} &ResNet-101~\cite{resnet101}& 69.9& 65.8 & - &-    \\
    UVTR~\cite{uvtr} & \ding{52} &ResNet-101~\cite{resnet101}& 70.2& 65.4 & - &-    \\
    GraphAlign~\cite{graphalign} & \ding{52} &ResNet-101~\cite{resnet101}& 70.6&  66.5  & - &  -  \\
    AutoAlignV2~\cite{autoalignv2} & \ding{52} &CSPNet~\cite{cspnet}&71.2& 67.1  &-  & -   \\
    CMT~\cite{cmt} & \ding{52} & ResNet-50~\cite{resnet101}&70.8 &67.9  & - &   - \\
    FUTR3D~\cite{futr3d} & \ding{52} &ResNet-101~\cite{resnet101}& 68.3&64.5  & 42.9 &35.1    \\
    TransFusion~\cite{transfusion} & \ding{52}&DLA34~\cite{dla34}& 70.2&  66.3 &  17.9&11.6     \\
    BEVFusion (MIT)~\cite{bevfusion} & \ding{52}&SwinT~\cite{swin}& 71.4&  68.5 & 7.1& 0.5   \\
    BEVFusion\dag(PKU)~\cite{bevfusion1} & \ding{52}&Dual-Swin-T~\cite{dualswint}& 70.4&  66.9 & 20.3& 14.2    \\
    MetaBEV~\cite{metabev} & \ding{52}& SwinT~\cite{swin}& 71.5& 68.0  &42.6&39.0   \\
    UniBEV~\cite{unibev} & \ding{52}&ResNet-101~\cite{resnet101}& 68.5&64.2   & 42.4&35.0   \\
    \midrule
    \rowcolor{gray!15}
    TransFusion~\cite{transfusion} & \ding{56}&DLA34~\cite{dla34}& 31.4& 26.9    &18.4&13.9  \\
    \rowcolor{gray!15}
    BEVFusion~\cite{bevfusion1} & \ding{56}&Dual-Swin-T~\cite{dualswint}&37.2& 32.4   &24.2& 17.3   \\
    \rowcolor{gray!15}
    VCD-E~\cite{vcd} & \ding{56}  & ConvNext-B~\cite{convnext} &71.1 &  67.7  & - &  -  \\
    \rowcolor{gray!15}
    \rowcolor{lgray}
    \textbf{LiteFusion-B} & \ding{56}&ResNet-101~\cite{resnet101}& 68.9&62.3 &44.3  &36.1 \\
    \rowcolor{lgray}
    \textbf{LiteFusion-L} & \ding{56}&ConvNext-B~\cite{convnext}&\textbf{72.9}  &\textbf{67.8} &\textbf{46.6} & \textbf{39.2} \\
    
   \bottomrule[1pt]
   \end{tabular} %
  }
 
\end{table*}

\subsection{Dataset and Implementation}
\label{sec:detail}
\noindent\textbf{Dateset.}
Our method was extensively evaluated using the nuScenes dataset, which is a large-scale, multi-modal dataset  for 3D object detection and map segmentation. The dataset includes a total of 1,000 driving sequences, divided into 700 for training, 150 for validation, and 150 for testing. Each sequence features data from six surrounding cameras that provide a complete 360-degree field of view, along with a 32-beam LiDAR system.

\noindent\textbf{Evaluation metrics.}
We use mean Average Precision (mAP) and NuScenes Detection Score (NDS) as our main evaluation metrics for comparison. The final mAP is calculated by averaging precision values across four distance thresholds for ten different classes. The NDS metric provides a weighted average of the mAP along with additional attribute metrics, which include translation, scale, orientation, velocity, and other box attributes.

\noindent\textbf{Implementation details.}
We use BEVFormer~\cite{bevformer} as our baseline camera-based detector, which employs ResNet-101\cite{resnet101} as the image feature extraction backbone. LiteFusion-S and LiteFusion-B are trained based on the BEVFormer-Small and BEVFormer-Base versions, respectively, to evaluate the performance of our framework under these configurations. The encoder layers follow the conventional structure of transformers~\cite{transformer}. Our codebase is on the basis of MMDetection3D~\cite{mmdet3d}, and the main experiments are conducted using 4 NVIDIA A800 GPUs. 
We mainly follow BEVFormer’s experimental setup for fair comparison, ensuring that our method is easily applicable to various foundational models.

For LiDAR input, we set the perception range to [-51.2 m, 51.2 m] on the X and Y axes, and [-5 m, 3 m] on the Z axis. To process the LiDAR data, we use a simple voxelization scheme to structure points, which are then projected onto the BEV plane along the z-axis. The voxel size is set to [0.23 m, 0.23 m, 8 m]. We train LiteFusion in an end-to-end manner with only one stage. 
Additionally, to ensure a fair comparison with other fusion methods, we use a finer voxel size of [0.075 m, 0.075 m, 0.2 m] and obtain a larger backbone ConvNext-B~\cite{convnext} for LiteFusion, referred to as LiteFusion-L. LiteFusion-L is trained on 8 NVIDIA A800 GPUs.

\subsection{Performance on 3D Object Detection} 
In this section, we emphasize the practical significance of LiteFusion from three important perspectives: ease of deployment, robustness, and scalability. Our demonstration establishes a baseline that outlines the challenges arising from the absence of a sparse convolution backbone for 3D feature extraction, as well as the impacts of LiDAR sensor corruptions, as illustrated in Table~\ref{tab:main}.

\vspace{8pt}
\noindent\textbf{High performance \& easy deployment.}
Previous fusion-based methods heavily rely on a 3D sparse backbone to extract point features, which necessitates complex, customized implementations and optimizations for deployment on various hardware. However, simply removing the 3D backbone leads to a significant performance drop, as shown in the gray rows of Table~\ref{tab:main}. This is mainly caused by the independent feature extraction processes used for both modalities. 
In contrast, LiteFusion is built entirely on canonical, well-optimized operators, avoiding the use of a 3D backbone. With this configuration, LiteFusion-B achieves state-of-the-art performance while maintaining broad compatibility and deployment-friendliness across diverse hardware platforms.
To further validate the effectiveness of our framework, we developed LiteFusion-L, which incorporates finer-grained point cloud mapping and a larger backbone. Despite not relying on a 3D sparse convolution backbone, LiteFusion-L achieves an impressive 72.9\% NDS and 67.8\% mAP, comparable to state-of-the-art methods. These results highlight that replacing the image backbone with a more powerful one, such as ConvNeXt-B, can further enhance performance. Notably, LiteFusion-L outperforms VCD-E~\cite{vcd}, which uses a long temporal sequence of 9 frames during training, by delivering superior performance with a shorter sequence of 6 frames.
%

\noindent\textbf{Robustness.}
Due to the explicit LiDAR branch providing rich geometrical information, previous multi-modal methods experience significant performance drops when LiDAR input is missing. In such cases, these methods default to zero-initialized features, as illustrated in the last column of Table~\ref{tab:main}. For example, approaches like BEVFusion~\cite{bevfusion1} and TransFusion~\cite{transfusion} face substantial decline in performance, with NDS reductions of 50.1\% and 52.3\%, respectively, when LiDAR input is not available.
In contrast, LiteFusion-L achieves impressive results, with an NDS of 46.6\% and a mAP of 39.2\%. These results demonstrate the superior robustness of the proposed LiteFusion, even when compared to methods that are specifically designed for enhancing robustness~\cite{metabev,unibev}.

\noindent\textbf{Scalability assessment.}
As illustrated in Table~\ref{tab:para}, our method only increases the number of parameters by 1.2\% on the BEVFormer-small version and by 1.1\% on the base version. With these small increases, we achieve a significant improvement in performance: 
LiteFusion-S achieves a +22.1\% increase in mAP and a +20.1\% increase in NDS, while 
LiteFusion-B achieves a +20.4\% increase in mAP and a +19.7\% increase in NDS. 
Our work emphasizes the importance of minimal parameter increases and limited modifications to the original camera-based detector. To facilitate a smooth adaptation, we retain the training strategies, optimization processes, and hyperparameters of the foundational vision-based detector. By considering LiDAR as an auxiliary modality, LiteFusion reduces reliance on separate feature extraction across modalities, significantly advancing camera-based detectors without additional customization.
Moreover, the results in Table~\ref{tab:para} demonstrate that LiteFusion delivers significant performance improvements with minimal increases in memory usage, and running speed, attributed to its lightweight architectural design.

\begin{table*}[tb]
  \centering
  \caption{\textbf{Effect of introducing geometry integrators to BEVFormer.} For the small and base versions of BEVFormer, the introduction of the LiDAR geometry integrator entails only a minimal increase in the number of parameters and GPU memory yet significantly enhances the performance of the detector.}
  \label{tab:para}
  \small 
  \setlength{\tabcolsep}{15pt} 
  \renewcommand{\arraystretch}{1.2} 
  \resizebox{\textwidth}{!}{
  \begin{tabular}{ccccccc}
    \toprule[1pt]    \textbf{Methods}&\textbf{NDS$\uparrow$}&\textbf{mAP$\uparrow$}&\textbf{Parameter}&\textbf{Memory}&\textbf{Training speed}&\textbf{FPS}\\
    \midrule
    \midrule
    BEVFormer-small~\cite{bevformer}&45.1 &36.2&59.46M& 10.25G&2.1s/iter&4.8 \\
    \rowcolor{gray!15}
    LiteFusion-S &65.2 \gain{($+20.1$)}& 58.3 \gain{($+22.1$)} & 60.17M \gain{($+1.2\%$)}& 12.65G&2.8s/iter&5.3\\
    
    \midrule
    BEVFormer-base~\cite{bevformer} & 49.2 & 41.9 &69.03M&27.83G&3.2s/iter&3.9 \\
    \rowcolor{gray!15}
    LiteFusion-B & 68.9 \gain{($+19.7$)}&62.3 \gain{($+20.4$)}    &69.81M \gain{($+1.1\%$)}&30.16G&3.7s/iter&4.6 \\
    \bottomrule[1pt]
  \end{tabular}}%
\end{table*}
\subsection{Performance on Different Vision-based Architectures} 
\begin{table}[!t]
    \centering
    \caption{\textbf{Performance comparison of different backbones and structures.} The results demonstrate that LiteFusion consistently achieves performance improvements across various architectures.}
    \label{tab:performance}
    \small
    \setlength{\tabcolsep}{12pt} 
    \renewcommand{\arraystretch}{1.2} 
    \resizebox{\linewidth}{!}{
    \begin{tabular}{cccc}
        \toprule
        \textbf{Methods} & \textbf{2D Backbone} & \textbf{NDS$\uparrow$ } & \textbf{mAP$\uparrow$ } \\
        \midrule
        \midrule
        BEVFormer~\cite{bevformer} & V2-99 & 55.3 & 47.2 \\
        \rowcolor{gray!15}
        +LiteFusion & V2-99 & 69.5 & 63.4 \\
        \midrule
        BEVFusion-C~\cite{bevfusion1} & Dual-Swin-T & 31.1 & 22.9 \\
        \rowcolor{gray!15}
        +LiteFusion & Dual-Swin-T & 57.6 & 50.2 \\
        \midrule
        BEVFormer v2~\cite{bevformerv2} & ResNet-101 & 45.1 & 37.4 \\
        \rowcolor{gray!15}
        +LiteFusion & ResNet-101 & 64.5 & 58.9 \\
        \midrule
        Sparse4D~\cite{lin2023sparse4d} & ResNet-50 & 55.8 & 45.2 \\
        \rowcolor{gray!15}
        +LiteFusion & ResNet-50 & 67.2 & 60.4 \\
        \bottomrule
    \end{tabular}
    }
\end{table}

We present additional results using various image backbones (V2-99, Dual-Swin-T, ResNet101, ResNet50) and architectures (BEVFusion, BEVFormer v2 and Sparse4D). BEVFormer v2 differs from its predecessor by introducing a new temporal encoder and a perspective 3D architecture. The experimental results, as shown in Table~\ref{tab:performance}, consistently demonstrate significant performance improvements with LiteFusion across these frameworks, highlighting our approach's adaptability to different architectures and configurations. 

LiteFusion has a simple design and does not incorporate additional network architectures and training schemes, as evidenced by the following:

\begin{enumerate}[label=(\roman*)]
    \item \textbf{Model structure:} LiteFusion preserves all model parameters of the vision-based model, extracting features from various layers to combine with point cloud data. This method does not alter the internal architecture of the layers and is not restricted to specific network structures.
    
    \item \textbf{Tuning settings:} LiteFusion retains the original structure and hyperparameter settings of the base detector, including the learning rate, training epochs, and optimizer choices. This straightforward design and execution allow LiteFusion to be easily adapted to different foundational detectors without requiring complex training techniques.
\end{enumerate}

\subsection{Qualitative Results}
\begin{figure*}[ht]
  \centering
  \includegraphics[width=\linewidth]{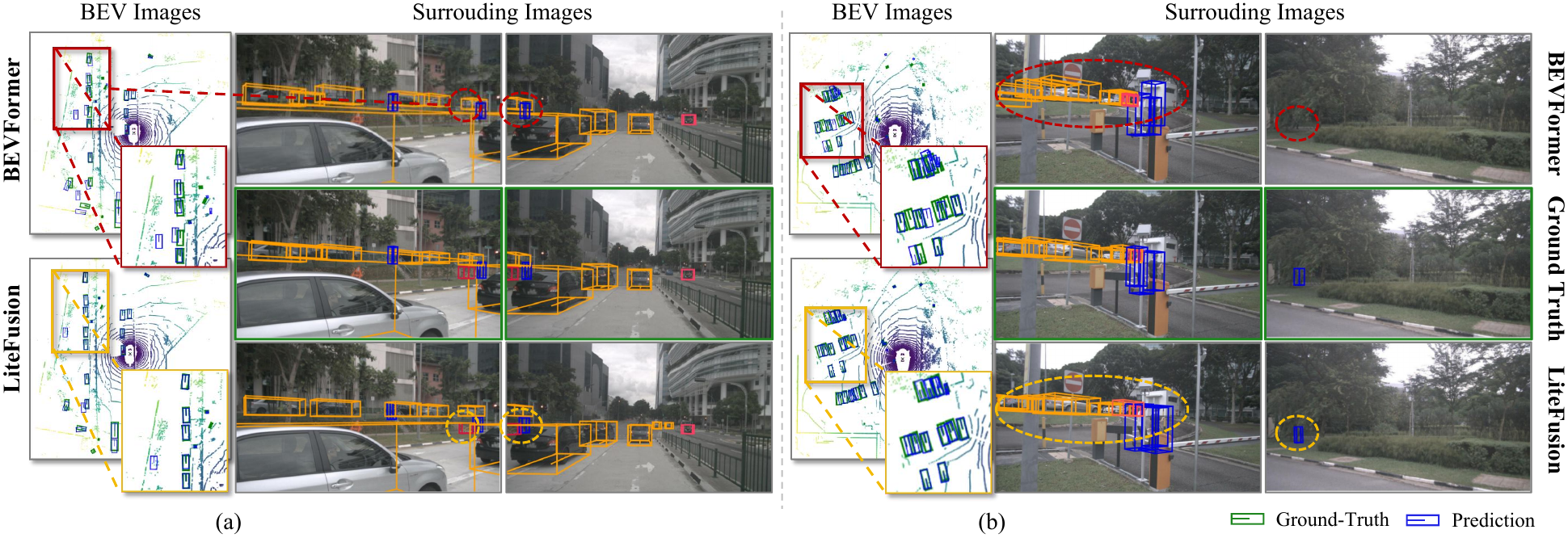}
  \caption{\textbf{Visualization results of LiteFusion.} The ground truth and predictions on the BEV plane are drawn in green and blue rectangles, respectively. LiteFusion obtains a more accurate detection result than the original BEVFormer with minimal adjustment.
  }
  \label{fig:visual}
\end{figure*}

In Fig.~\ref{fig:visual}, we report the 3D object detection results on the nuScenes dataset, comparing our proposed model, LiteFusion-B, with BEVFormer. 
Fig.~\ref{fig:visual}(a) shows LiteFusion’s capability to maintain accurate detection rates while ensuring precise bounding box localization. The first row, highlighted with red circles, reveals that BEVFormer struggles to detect small targets, whereas LiteFusion consistently delivers more accurate results.
Moreover, Fig.~\ref{fig:visual}(b) highlights LiteFusion’s superior capability in 3D spatial geometry perception. In regions with dense targets, LiteFusion achieves more precise bounding box dimensions and localization compared to BEVFormer.
LiteFusion outperforms BEVFormer by progressively integrating spatial geometry from LiDAR with image features. This integration enhances cross-modal alignment and improves depth perception. 
By effectively utilizing spatial geometry derived from point clouds, LiteFusion achieves more accurate detection, especially in challenging situations like identifying small targets and navigating densely populated areas.

\subsection{Performance on Data Scalability} 

To further validate the data scalability of our method, we train LiteFusion-S with different portions of camera-only and multi-modal data. 
In Table \ref{tab:results}, we have the following observations:

\begin{itemize}
    \item[(i)] Comparing the second and third rows, when the camera data increases from 50\% to 100\%, mAP dramatically increases from 38.4\% to 45.3\%, showing LiteFusion's favorable scalability on camera data.
    \item[(ii)] Comparing the third and last rows, when the multi-modal data increases from 25\% to 100\%, mAP increases from 45.4\% to 58.3\%, affirming LiteFusion's favorable scalability on multi-modal data.
\end{itemize}

\begin{table}[!t]
    \centering
    \caption{\textbf{Performance across varying volumes of multimodal data.} LiteFusion exhibits good scalability with both camera data and multi-modal data.}
    \label{tab:results}
    \setlength{\tabcolsep}{8pt} 
    \resizebox{\linewidth}{!}{
    \begin{tabular}{ccccc}
        \toprule[1pt]
        \textbf{Method} & \textbf{Camera} & \textbf{Multi-modal} & \textbf{NDS$\uparrow$ }& \textbf{mAP$\uparrow$ }  \\
        \midrule
        \midrule
        LiteFusion & 50\% & 25\% & 42.1& 38.4  \\
        LiteFusion & 100\% & 25\% & 53.7& 45.3  \\
        LiteFusion & 100\% & 100\% & 65.2& 58.3  \\
        \bottomrule[1pt]
    \end{tabular}
    }
\end{table}

\subsection{Parameter Efficiency of Quaternion Layers} 

LiteFusion introduces cross-modal mapping into every layer of the vision-based model, but these mappings are parameter-efficient, increasing the total parameters by only 1.1\%. This efficiency is achieved by integrating information from both modalities using a quaternion layer. Alternative designs, such as vanilla MLPs and concatenation, are also evaluated. The experimental results in Table \ref{tab:GAEfusion} demonstrate that quaternion representation achieves superb performance while using only a quarter of the parameters typically required by vanilla MLPs.

\begin{table}[!t]
    \centering
    \caption{\textbf{Performance comparison of fusion operations in DAE.} Quaternion Layer Achieves Significant Improvements with Fewer Parameters.}
    \label{tab:GAEfusion}
    \setlength{\tabcolsep}{8pt} 
    \resizebox{\linewidth}{!}{
    \begin{tabular}{lccc}
        \toprule[1pt]
        \textbf{Ops. in DAE} & \textbf{NDS$\uparrow$}& \textbf{mAP$\uparrow$}  & \textbf{Param. (Rel. to MLP)} \\
        \midrule
        \midrule
        Concatenation             & 62.4& 55.7  & $+0$ \\
        MLP                       & 63.6 & 56.9 & $+100\%$\\
        \rowcolor{gray!15}
        Quaternion Layer          & 65.2& 58.3  & $+25\%$\\
        \bottomrule[1pt]
    \end{tabular}}
\end{table}

In summary, since our goal is to empower a vision-based detector at the lowest possible cost, 
the main contribution of LiteFusion is not in achieving superior accuracy for a multi-modal fusion detector. Instead, we focus on addressing a different issue: developing a multi-modal detector that is better suited for practical applications. Specifically,

\begin{enumerate}[label=(\roman*)]
    \item We aim to avoid the use of a 3D backbone, which is commonly utilized in almost all fusion methods. The sparse convolutions required for 3D backbones need customized implementation and optimization to work effectively on various hardware platforms (e.g., NPUs or FPGAs), limiting their use in real-world production scenarios.
    \item We focus on integrating LiDAR data with visual data in a plug-and-play manner. Instead of processing the point cloud from LiDAR through explicit feature extraction methods like in earlier dual-stream approaches, we treat it as a supplementary geometric reference to be embedded within image features. This strategy enables vision-based models to leverage cross-modal information effectively.
\end{enumerate}

\subsection{Ablation Studies} 
\subsubsection{Effect of Proposed Components}

Our ablation studies, presented in Table~\ref{tab:ablation}, demonstrate that combining GAE with DAE significantly improves performance compared to using DAE alone.
The complete integration of components leads to a significant improvement in the final mAP and NDS, increasing from 36.2\% to 58.3\% and from 45.1\% to 65.2\%, respectively. We suggest that while the DAE provides depth information from LiDAR projections to support 2D feature extraction, this additional information is not fully utilized by the BEV encoder when used independently. When both modules are used together, the geometric information derived from LiDAR serves as prior knowledge, thereby unlocking the depth perception potential of the 2D feature backbone. This comprehensive setup demonstrates the synergistic effect of depth and spatial geometry context information, highlighting LiteFusion's superiority in enhancing 3D object detection.

\begin{table}[!t]  
    \caption{\textbf{Ablation studies on proposed components.} We explored the effects of DAE, GAE for encoder, and GAE for decoder individually and in combination. The results show that integrating these components amplifies their respective benefits.}
    \label{tab:ablation}
    \centering
    \small 
    \setlength{\tabcolsep}{8pt} 
    \resizebox{\linewidth}{!}{
    \begin{tabular}{ccccc} 
			\toprule[1pt]
             \multicolumn{3}{c}{\textbf{Components of LiteFusion}}  &\multirow{2}{*}{\textbf{NDS$\uparrow$}}&\multirow{2}{*}{\textbf{mAP$\uparrow$}}\\
             \cmidrule(lr){1-3} 
             {\textbf{DAE}}&{\textbf{GAE-Enc}}&{\textbf{GAE-Dec}}&&\\
             \midrule
             \midrule
             \ding{56}&\ding{56}&\ding{56}& {45.1 }&{36.2}\\  
             \midrule
			\ding{52} &\ding{56}&\ding{56}  & $+ 3.9$ & $+ 4.2$      \\
             \ding{56}&\ding{52}&\ding{56}  & $+9.7$& $+ 10.2$  \\
             \ding{56}&\ding{56} &\ding{52}  &$+5.2$& $+ 5.8$  \\
            \cdashline{1-5}[2pt/2pt]
			\ding{56}&\ding{52} & \ding{52}& $+ 15.3$& $+ 16.4$  \\ 
            \ding{52}&\ding{52} & \ding{56}& $+ 16.2$& $+ 17.1$  \\ 
            \midrule
            \ding{52} & \ding{52} & \ding{52}  &  \textbf{{65.2 \gain{($+20.1$)}}}&  \textbf{{58.3 \gain{($+22.1$)}}}  \\ 
			\bottomrule[1pt]
	\end{tabular}
    }
\end{table}

\subsubsection{Efficacy of Progressive Response Framework}
\begin{figure*}[tb]
  \centering
  \includegraphics[width=\linewidth]{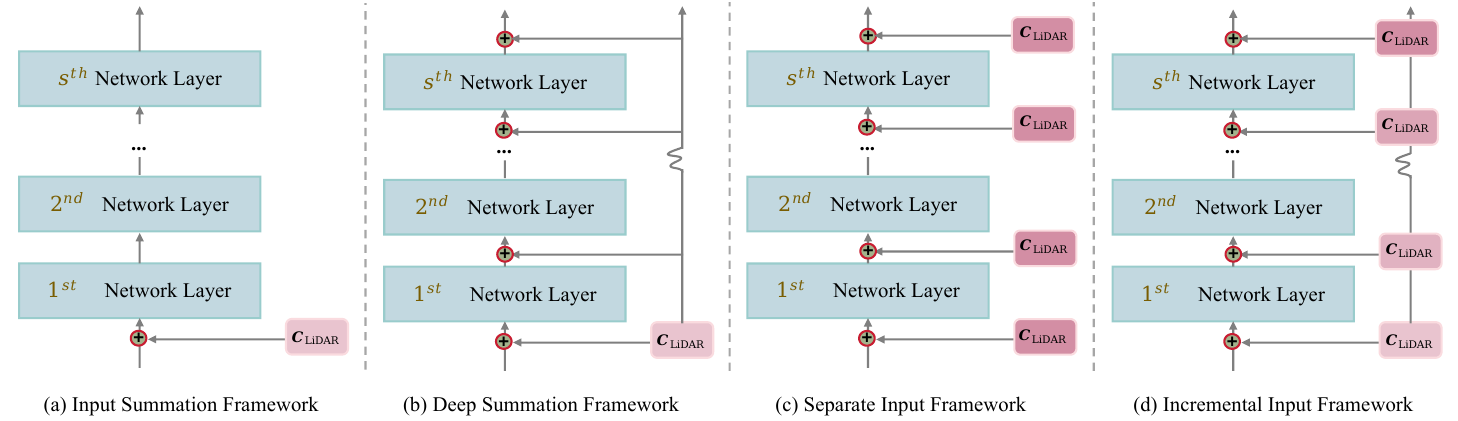}
  \caption{\textbf{Schematic diagrams of the four different progressive fusion framework setups.} The geometry embedding features $\bm{C}_\text{LiDAR}$ are fed into the camera-based model in four ways to explore the effectiveness of the proposed progressive response framework. 
  }
  \label{fig:experiment}
\end{figure*}
We analyze the performance enhancements achieved by adopting our designed progressive response framework for integrating point cloud information. 
As shown in Fig.~\ref{fig:experiment}, we select four alternative configurations, which differ in how the geometry embedding features are fed into the camera-based model. Their experimental results are presented in Table~\ref{tab:framework}.
Our analysis shows that integrating geometry embedding from point cloud projections into the base model—whether at the input stage or in later layers, as illustrated in Fig. \ref{fig:experiment}(a) and (b)—results in only minimal performance improvements. This demonstrates that without an effective integration structure, the information gained from directly inputting point cloud data is limited, underscoring the contributions of our proposed framework.
Conversely, the introduction of a geometric information integrate module, as demonstrated in Fig.~\ref{fig:experiment}(c), markedly elevates model performance, yielding a significant rise of $+17.4\%$ in mAP and $+16.0\%$ in NDS.
The peak performance is attained through a progressive fusion strategy, detailed in Fig.~\ref{fig:experiment}(d).

\begin{table}[!t]
   \centering
   \caption{\textbf{Ablation studies on fusion framework.} The experimental results reveal that simply feeding the point cloud into the camera-based framework without the 3D feature extraction stage yields very limited outcomes. The point cloud information is enhanced by the proposed progressive response strategy, resulting in substantial improvements in performance.}
   \label{tab:framework}
   \small
   \setlength{\tabcolsep}{16pt} 
   \renewcommand{\arraystretch}{1.2} 
   \resizebox{\linewidth}{!}{
   \begin{tabular}{lcc}
     \toprule[1pt]    
     \textbf{Methods} & \textbf{NDS$\uparrow$}& \textbf{mAP$\uparrow$}  \\
     \midrule
     \midrule
     Baseline            & 45.1 & 36.2 \\
     \midrule
     Input Summation     & 47.8 \gain{($+2.7$)}& 37.1 \gain{($+0.9$)}  \\
     Deep Summation      & 49.6 \gain{($+4.5$)} & 39.2 \gain{($+3.0$)} \\
     Separate Input     & 61.1 \gain{($+16.0$)}& 53.6 \gain{($+17.4$)}  \\
     \rowcolor{gray!15}
     Progressive Input   & 65.2 \gain{($+20.1$)}& 58.3 \gain{($+22.1$)}  \\
     \bottomrule[1pt]
   \end{tabular}%
   }
 \end{table}

\subsubsection{Effect of Quaternion Feature Adaptation Module}

\begin{table}
	\centering
    \caption{\textbf{Comparison of the effects of introducing quaternion feature adaptation mechanism.} Our method performs well over the vanilla methods due to successful cross-modal mining with quaternion space modulation.}
    \label{tab:quaternion}
    \small
    \setlength{\tabcolsep}{24pt} 
    \renewcommand{\arraystretch}{1.2} 
    \resizebox{\linewidth}{!}{
	\begin{tabular}{lcc}
    \toprule[1pt]    
    \textbf{Methods}&\textbf{NDS$\uparrow$}&\textbf{mAP$\uparrow$}\\
    \midrule
    \midrule
    Baseline&45.1&36.2  \\
    \midrule
    Ours w/o Qua-FA&62.4 &55.7\\
    \rowcolor{gray!15}
    Ours &\textbf{65.2}&\textbf{58.3}\\
    \bottomrule[1pt]
  \end{tabular}%
  }
\end{table}
 We conducted further experiments as shown in Table~\ref{tab:quaternion} to evaluate the efficacy of the proposed Quaternion Feature Adaptation (Qua-FA). 
 The results indicate that the quaternion network efficiently models the orthogonal image-LiDAR relationships, which leads to further improved performance.

\begin{figure}
       \centering
       \includegraphics[width=0.95\linewidth]{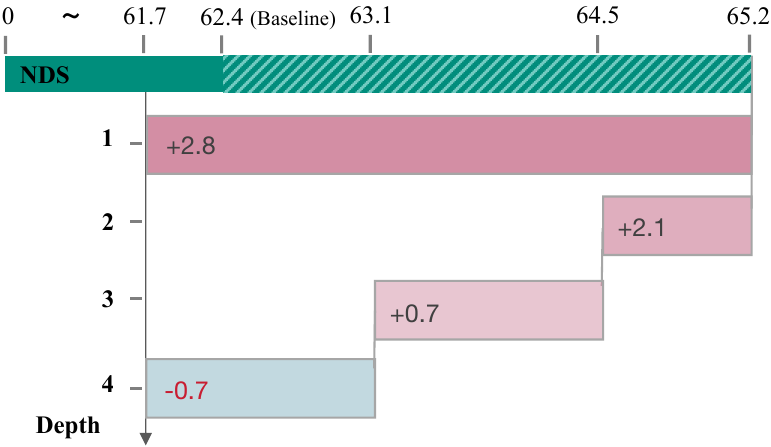} 
       \caption{\textbf{Comparison of Qua-FA insertion depths.} The green bar indicates rising NDS from left to right, with the critical point being the baseline. Pink bars show increases in NDS, while blue bars indicate undesired decreases.}
        \label{fig:qnet}
\end{figure}

We also investigated how the depth at which Qua-FA is introduced affects the results. The findings shown in Fig.~\ref{fig:qnet} indicate that applying Qua-FA solely at the first layer, where it aligns the two modalities in a quaternion space, produces the best outcomes. In contrast, placing Qua-FA in deeper layers disrupts the interaction between modalities in the feature space. 

Essentially, Qua-FA provides an orthogonal spatial prior to the beginning of the feature extraction process. This spatial prior helps guide the network in aligning information, directing the subsequent feature fusion. However, applying this orthogonal constraint closer to the output can hinder the fusion of modalities, resulting in an undesired decrease in performance.

\subsection{Further Discussion}
\label{sec:further}

\subsubsection{Effect of Framework Depth}
\begin{table*}[ht]
    \centering
    \begin{minipage}{0.32\textwidth}
        \centering
        \caption{Different insertion depth of \textbf{DAE} for \textbf{image backbone}.}
        \label{tab:f1}
        \small
        \setlength{\tabcolsep}{16pt}
        \renewcommand{\arraystretch}{1.2}
        \resizebox{\textwidth}{!}{
            \begin{tabular}{ccc}
                \toprule[1pt]
                \textbf{Depth} & \textbf{NDS$\uparrow$}& \textbf{mAP$\uparrow$}  \\
                \midrule
                \midrule
                1 & 62.1&55.2 \\
                2 &63.7&56.9 \\
                \rowcolor{gray!15}
                4 &65.2 &58.3 \\
                \bottomrule[1pt]
            \end{tabular}%
        }
    \end{minipage}%
    \hfill
    \begin{minipage}{0.32\textwidth}
        \centering
        \caption{Different insertion depth of \textbf{GAE} for BEV \textbf{encoder}.}
        \label{tab:f2}
        \small
        \setlength{\tabcolsep}{16pt}
        \renewcommand{\arraystretch}{1.2}
        \resizebox{\textwidth}{!}{
            \begin{tabular}{lcc}
                \toprule[1pt]
                \textbf{Depth} & \textbf{NDS$\uparrow$}& \textbf{mAP$\uparrow$}  \\
                \midrule
                \midrule
                Input &57.9&51.2  \\
                \rowcolor{gray!15}
                Interlace & 64.7&58.1 \\
                \rowcolor{gray!15}
                All &65.2&58.3  \\
                \bottomrule[1pt]
            \end{tabular}%
        }
    \end{minipage}%
    \hfill
    \begin{minipage}{0.32\textwidth}
        \centering
        \caption{Different insertion depth of \textbf{GAE} for BEV \textbf{decoder}.}
        \label{tab:f3}
        \small
        \setlength{\tabcolsep}{16pt}
        \renewcommand{\arraystretch}{1.2}
        \resizebox{\textwidth}{!}{
            \begin{tabular}{lcc}
                \toprule[1pt]
                \textbf{Depth} & \textbf{NDS$\uparrow$}& \textbf{mAP$\uparrow$}  \\
                \midrule
                \midrule
                Input &62.1& 56.5 \\
                Interlace &64.4 &57.1 \\
                \rowcolor{gray!15}
                All &65.2&58.3  \\
                \bottomrule[1pt]
            \end{tabular}%
        }
    \end{minipage}
\end{table*}
We examined the impact of integrating integrators at various depths within a progressive response framework on model performance. Tables \ref{tab:f1}, \ref{tab:f2}, and \ref{tab:f3} illustrate the effects across three network components: DAE for the image backbone, GAE for the encoder, and GAE for the decoder. Our findings indicate a positive correlation between the depth of integrator integration and performance enhancement, confirming the effectiveness of the progressive response framework in utilizing LiDAR information.

In Table \ref{tab:f2}, we observe that inserting integrators in every other layer of the BEV encoder does not lead to a significant decrease in performance. This finding confirms that intermittent insertion is a valid strategy for reducing parameters within our framework. It suggests that a non-continuous use of the integrator can still allow the network to effectively utilize LiDAR data. This insight supports a strategic approach that balances model compactness with the efficient use of multi-modal sensory information.

\subsubsection{Effect of Hidden Space Dimensions}

\begin{figure}[t]
  \centering
  \includegraphics[width=0.95\linewidth]{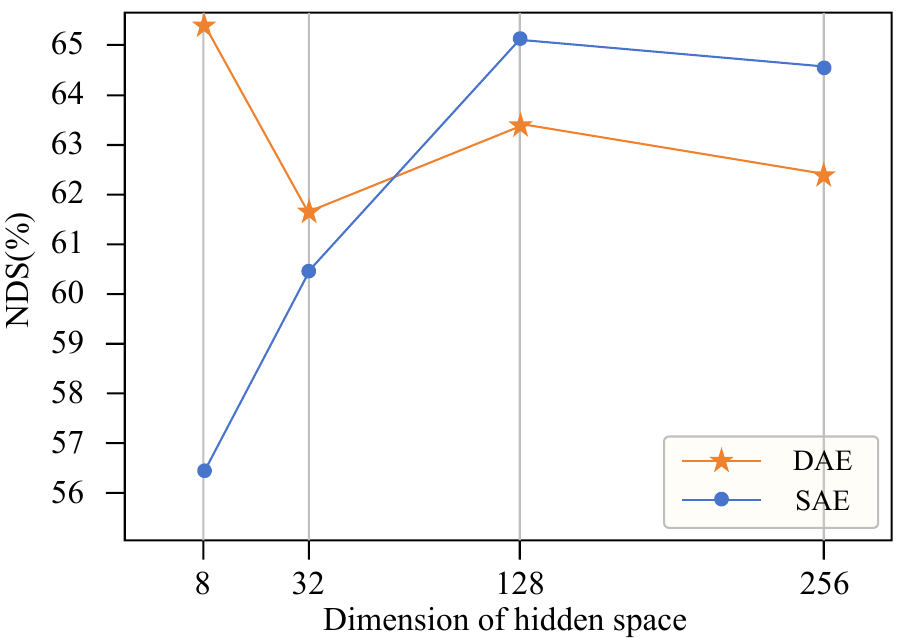}
  \caption{\textbf{Results on different dimensions of hidden space in DAE and GAE.} Optimal Performance Achieved at 8 Dimensions for DAE and 128 Dimensions for GAE.}

  \label{fig:dimension}
\end{figure}

LiteFusion designs the LiDAR geometry integrator using a strategy of ``dimension reduction, alignment, and expansion.'' In Fig.~\ref{fig:dimension}, we evaluate the effects of different hidden space dimensions during the dimension reduction process. Notably, for the DAE, setting the dimension to 8 yields better performance compared to using higher dimensions. In contrast, the GAE achieves optimal performance at 128 dimensions. This indicates that lower dimensions may be more effective for DAE in capturing essential features, while GAE benefits from higher-dimensional space, allowing it to encapsulate more complex 3D geometric information.

\subsubsection{Effect of Quaternion Mapping in BEV and PV Spaces}
Our experiments show that using quaternions in the BEV space does not offer significant advantages. Table~\ref{tab:quaternion_comparison} illustrates that the benefits of employing quaternion mapping are minimal compared to simple concatenation.
\begin{table}
	\centering
    \caption{\textbf{Performance comparison of fusion operations in GAE.} Compared to simple concatenation, the use of quaternion layers in GAE shows a modest but limited improvement. }
    \setlength{\tabcolsep}{8pt} 
    \label{tab:quaternion_comparison}
    \resizebox{\linewidth}{!}{
	\begin{tabular}{lccc}
    \toprule[0.5pt]    
    \textbf{Ops. in GAE} & \textbf{NDS$\uparrow$}& \textbf{mAP$\uparrow$} & \textbf{Param. (Rel. to MLP)}\\
    \midrule
    \midrule
    Linear Layer & 63.8& 56.4  &$+100\%$\\
    \rowcolor{gray!15}
    Quaternion Layer & 64.6& 57.4 & $+25\%$\\
    \rowcolor{gray!15}
    Concatenation & 64.3& 57.2 & $+0\%$\\
    \bottomrule[1pt]
  \end{tabular}%
  }
\end{table}

In general, we speculate that the limited improvements observed with the quaternion layer in BEV space can be attributed to the fact that BEV features do not need to encode 3D information; this information is already represented in the voxel indices where the features are located. Consequently, fusing data from both homogeneous modalities information can be quite straightforward, often achieved simply by averaging the features within the same voxels. This reduces the need for a quaternion layer to enhance 3D information encoding. 
In contrast, quaternion layers perform well in PV space for two main reasons:
\begin{enumerate}[label=(\roman*)]
\item PV images do not inherently encode 3D information, and quaternion mappings can effectively address perspective differences between 2D image data and 3D LiDAR inputs by representing rotations and three-dimensional transformations in point clouds. 
\item There is a significant disparity between the two modalities, and quaternion layers are beneficial in mitigating this disparity.
\end{enumerate}

\subsubsection{Effect of Quaternion Elements Assignment}

We further explore the impact of assigning positions to LiDAR and image modalities within quaternion space. Specifically, we compare two arrangements: one where \( \bm{Q}_h = \bm{F}_{\text{img}}^r1 + \bm{C}_{\text{depth}}^ri + 0j + 0k \) allocates the LiDAR features to the imaginary part of the quaternion, and another where \( \bm{Q}_h = \bm{C}_{\text{depth}}^r1 + \bm{F}_{\text{img}}^ri + 0j + 0k \) places the LiDAR features on the real axis.
The experimental results, summarized in Table~\ref{tab:quari}, illustrate the outcomes of these different assignments. A notable improvement is observed when the depth information is assigned to the imaginary i-axis, with the mAP increasing to 58.3\% and the NDS reaching 65.2\%. Our findings indicate that positioning LiDAR features on the imaginary components—typically associated with 3D spatial dimensions—yields significant improvements compared to assignments on the real axis. This suggests that the multidimensional nature of LiDAR data may resonate more effectively with the spatially expressive imaginary axis of the quaternion space, thereby optimizing the fusion process within the LiteFusion framework.
 \begin{table}[t]
  \centering
  \caption{\textbf{Impact of LiDAR axis allocation in quaternion space.} Positioning LiDAR on the imaginary axis shows a clear advantage.}
  \label{tab:quari}
  \small 
  \setlength{\tabcolsep}{22pt} 
  \renewcommand{\arraystretch}{1.2} 
  \resizebox{\linewidth}{!}{
  \begin{tabular}{lcc}
    \toprule[1pt]    
\textbf{Methods}&\textbf{NDS$\uparrow$}&\textbf{mAP$\uparrow$}\\
    \midrule
    \midrule
    $\bm{C}_\text{depth}$ on the r-axis&{63.1}&{56.4}  \\
    \rowcolor{gray!15}
    $\bm{C}_\text{depth}$ on the i-axis &{65.2}&{58.3}\\
    \bottomrule[1pt]
  \end{tabular}%
  }
\end{table}

\subsubsection{Extension Visualization}
\begin{figure*}[ht]
  \centering
  \includegraphics[width=\linewidth]{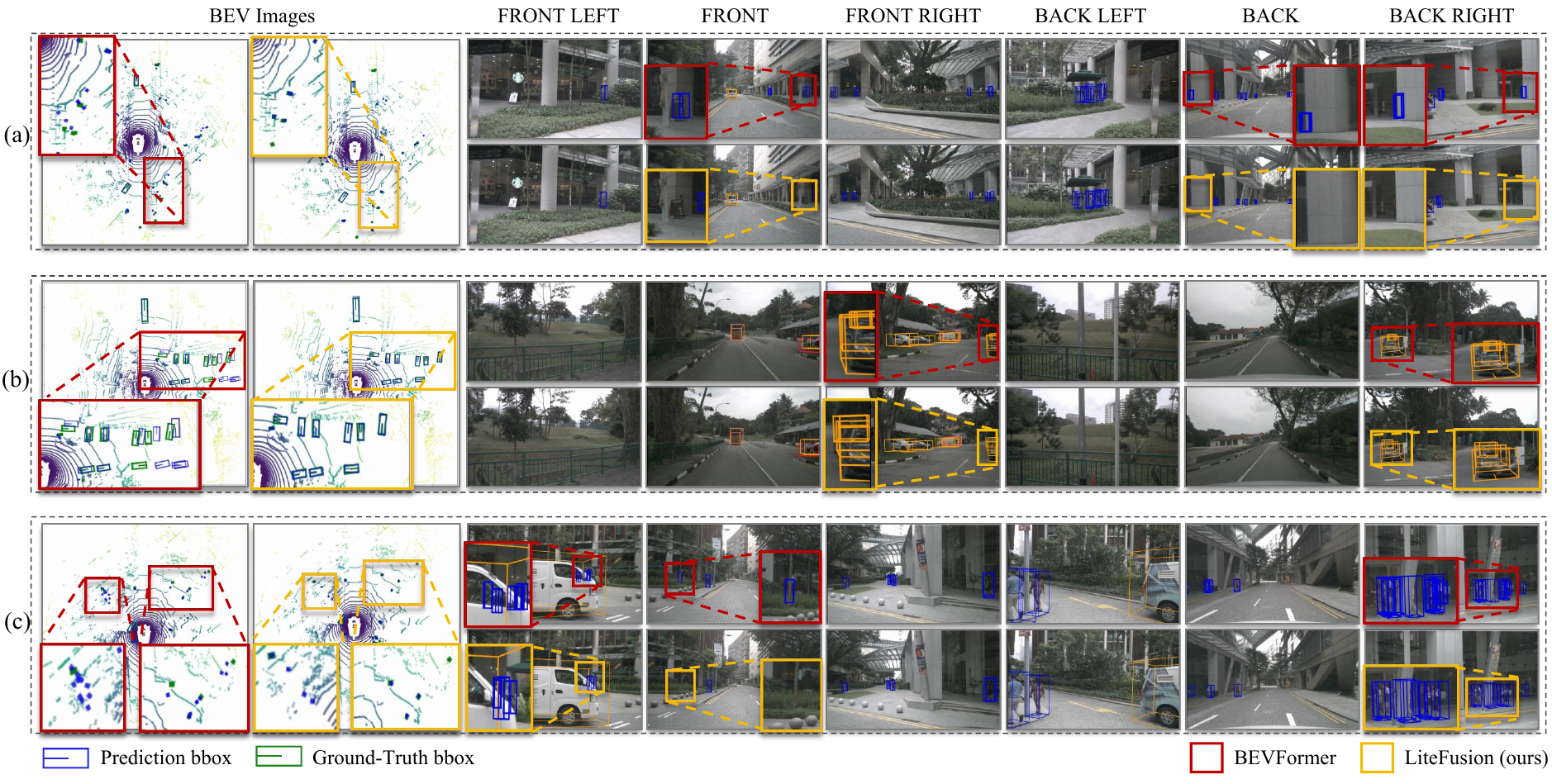}
  \caption{\textbf{Extended visualization results of LiteFusion in various challenging scenarios.} (a) Dense small object detection. (b) Scenarios with non-uniform object distribution. (c) Extremely dense environments. The ground truth and predictions on the BEV plane are drawn in green and blue rectangles, respectively. }
  \label{fig:main_visual2}
\end{figure*}

We present additional visualization results across diverse scenarios. Figure \ref{fig:main_visual2}(a) highlights performance in dense small object detection. It demonstrates LiteFusion's ability to parse crowded scenes where traditional models may struggle due to the complex overlap and proximity of objects. Figure \ref{fig:main_visual2}(b) further illustrates LiteFusion’s capability to refine bounding boxes for precise object localization and size estimation, showcasing detailed object representation. Lastly, Figure \ref{fig:main_visual2}(c) contrasts the reduction of false positives between LiteFusion and BEVFormer. By leveraging the complementary strengths of LiDAR and camera data with minimal architectural adjustments, LiteFusion navigates dense object scenes with enhanced accuracy and improves object representations for better localization and classification.

\section{Conclusion}

In this paper, we introduce LiteFusion, a novel fusion framework designed to bridge the gap between vision-based and multi-modal 3D object detectors with minimal structural modifications. LiteFusion incorporates a set of meta-geometry integrators that progressively integrate LiDAR information into the camera-based detection pipeline within an incrementally responsive framework. 
By employing quaternion space mapping, LiteFusion generates additional spatial geometric information from LiDAR data, allowing for the efficient and mathematically coherent encoding of complex 3D spatial relationships and rotations. Our research demonstrates that it is possible to significantly enhance the performance of 3D object detectors without relying on complex LiDAR feature extraction processes. 
This fusion paradigm improves robustness and scalability for model deployment.

\bibliographystyle{IEEEtran}
\bibliography{main}

\vfill

\end{document}